\documentclass[conference]{IEEEtran}

\usepackage{amsmath,amssymb,amsfonts}
\usepackage{graphicx}
\usepackage{textcomp} 
\usepackage[table,dvipsnames]{xcolor} 
\usepackage{colortbl}
\usepackage{booktabs}  
\usepackage{array}     
\usepackage{listings}  
\usepackage{multirow}  
\usepackage{hyperref}  
\usepackage[misc,geometry]{ifsym}
\usepackage{orcidlink}
\definecolor{lightblue}{RGB}{220, 230, 241}
\definecolor{background}{RGB}{240,240,240}
\definecolor{pass}{RGB}{200,255,200} 
\definecolor{fail}{RGB}{255,200,200} 
\definecolor{deepblue}{rgb}{0,0,0.5}
\definecolor{deepred}{rgb}{0.6,0,0}
\definecolor{deepgreen}{rgb}{0,0.5,0}
\usepackage{array}
\usepackage{mathtools}
\makeatletter
\newcommand{\printfnsymbol}[1]{%
  \textsuperscript{\@fnsymbol{#1}}%
}
\makeatother
\DeclareFixedFont{\ttb}{T1}{txtt}{bx}{n}{7} 
\DeclareFixedFont{\ttm}{T1}{txtt}{m}{n}{7}  
\newcommand\pythonstyle{\lstset{
language=Python,
basicstyle=\fontsize{7}{9}\selectfont\ttfamily,
morekeywords={self},              
keywordstyle=\ttb\color{deepblue},
emph={MyClass,__init__},          
emphstyle=\ttb\color{deepred},    
stringstyle=\color{deepgreen},                       
showstringspaces=false, 
breaklines=true,
numberbychapter=false,
}}

\lstnewenvironment{python}[1][]
{
\pythonstyle
\lstset{#1}
}
{}
\begin{document}

\title{Generation of Programmatic Rules for Document Forgery Detection Using Large Language Models}

\author{\IEEEauthorblockN{1\textsuperscript{st} Valentin Schmidberger}
\IEEEauthorblockA{
            \textit{Stuttgart Media University}\\
            Stuttgart, Germany \\
            schmidberger@hdm-stuttgart.de
            }
\and  
\IEEEauthorblockN{1\textsuperscript{st} Manuel Eberhardinger}
\IEEEauthorblockA{
            \textit{Stuttgart Media University}\\
            Stuttgart, Germany \\
            eberhardinger@hdm-stuttgart.de
            }
\and
\IEEEauthorblockN{2\textsuperscript{nd} Setareh Maghsudi}
\IEEEauthorblockA{
            \textit{Ruhr-University Bochum}\\
            Bochum, Germany \\
            setareh.maghsudi@rub.de
            }  
\and
\IEEEauthorblockN{3\textsuperscript{rd} Johannes Maucher}
\IEEEauthorblockA{
            \textit{Stuttgart Media University}\\
            Stuttgart, Germany \\
            maucher@hdm-stuttgart.de
            }
}

\maketitle 

\begin{abstract}
Document forgery poses a growing threat to legal, economic, and governmental processes, requiring increasingly sophisticated verification mechanisms. One approach involves the use of plausibility checks, rule-based procedures that assess the correctness and internal consistency of data, to detect anomalies or signs of manipulation. Although these verification procedures are essential for ensuring data integrity, existing plausibility checks are manually implemented by software engineers, which is time-consuming. Recent advances in code generation with large language models (LLMs) offer new potential for automating and scaling the generation of these checks. However, adapting LLMs to the specific requirements of an unknown domain remains a significant challenge. This work investigates the extent to which LLMs, adapted on domain-specific code and data through different fine-tuning strategies, can generate rule-based plausibility checks for forgery detection on constrained hardware resources. We fine-tune open-source LLMs, Llama 3.1 8B and OpenCoder 8B, on structured datasets derived from real-world application scenarios and evaluate the generated plausibility checks on previously unseen forgery patterns. The results demonstrate that the models are capable of generating executable and effective verification procedures. This also highlights the potential of LLMs as scalable tools to support human decision-making in security-sensitive contexts where comprehensibility is required.

\end{abstract}

\begin{IEEEkeywords}
Deep Learning, Large Language Models, Domain-Specific Fine-Tuning, Document Forgery
\end{IEEEkeywords}

\section{Introduction}
\label{sec:intro}
The rapid advancement of digitization and globalization has led to a notable increase in document forgery, which poses major challenges for both the private and public sectors. The proliferation of digital tools and interconnected global markets facilitates the creation and dissemination of counterfeit documents, with reported cases rising by 15.9\% between 2019 and 2023 \cite{GroHeidi.}. This emphasizes the urgent need for innovative solutions.

Automated AI-based systems are increasingly crucial for detecting patterns and anomalies in document verification processes \cite{gupta2020forensic,takenaka2024classification} and also for automating procedures to comply with legal regulations \cite{eberhardinger2025anonymization}.  
The main objective of this work is the adaptation of pre-trained large language models (LLMs) to generate plausibility checks, represented as Python code, that support the detection of document forgeries. Plausibility checks are rule-based procedures that assess the correctness and internal consistency of data, e.g., whether the issuing date of the document is an official working day of the issuing country; another check verifies if the document number is unique in the entire database.

This involves leveraging different fine-tuning strategies to incorporate domain-specific knowledge and developing mechanisms to generalize across emerging forgery patterns. Furthermore, privacy concerns require the use of models that operate entirely on locally stored data, since closed-source models, such as GPT-4 \cite{OpenAI.3152023} from OpenAI, process user input externally, which conflicts with data protection guidelines in security sensitive contexts. Another prerequisite for the use of AI models in law enforcement is the need for comprehensibility of the prediction, which we circumvent by using LLMs to generate program code that is inherently interpretable and can therefore be checked for correctness by experts before these checks are used to support the decision-making processes of employees.

\begin{figure*}[t]
    \centering
    \includegraphics[width=\linewidth]{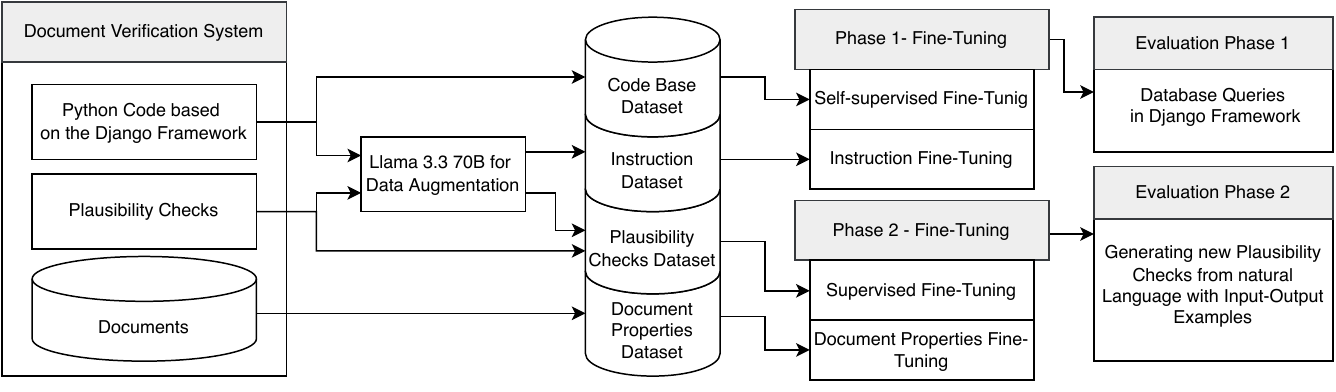}
    \caption{Overview of the methodology. We start with the existing document verification system, which consists of a Python code base, plausibility checks, and documents stored in the database. From this we create four datasets for fine-tuning and use Llama 3.3 70B for data augmentation. The fine-tuning is divided into two phases: The first phase focuses on fine-tuning the models on the code base dataset and instruction dataset, while the second phase involves further fine-tuning on the existing plausibility checks and document properties of the database. In the first phase, the models are evaluated on their ability to generate executable code for the document verification system, whereas in the second phase the generated plausibility checks are evaluated on their performance in detecting unseen document forgeries. }
    \label{fig:methodology}
\end{figure*}

Figure \ref{fig:methodology} shows an overview of the methodology. We start with the existing document verification system, which consists of a Python code base, plausibility checks, and documents stored in the database. From this we create four datasets for fine-tuning. The fine-tuning is divided into two phases: The first phase focuses on fine-tuning the models on the code base dataset and instruction dataset, while the second phase involves further fine-tuning on the existing plausibility checks and document properties of the database.
We use four LLMs from two model families, Llama 3.1 8B (base/instruct) \cite{Grattafiori.7312024} and OpenCoder 8B (base/instruct) \cite{Huang.1172024} and fine-tune them with Low-Rank Adaptation \cite{hu2022lora} on constrained hardware resources, namely a single Nvidia RTX 6000 48GB GPU. 
To augment existing data, Llama 3.3 70B is used to generate synthetic data following the method proposed in \cite{Grattafiori.7312024,Wang.12202022}. In the first phase, the models are evaluated on their ability to generate executable code for the document verification system, while in the second phase the generated plausibility checks are evaluated on their performance in detecting unseen document forgeries. 
Our code and data are part of proprietary software and therefore cannot be made available as open source. 

This paper is structured as follows: We start with related work in Sect. \ref{sec:related}, followed by the background of the document verification system that defines the context of this work (Sect. \ref{sec:background}). Afterwards, the methodology for creating the datasets for phase 1 (Sect. \ref{sec:phase1}) and phase 2 (Sect. \ref{sec:phase2}) is explained. Each phase includes an evaluation and discussion of the results, followed by an overall discussion of this study in Sect. \ref{sec:discussion}. We conclude the paper in Sect. \ref{sec:conclusion}.

\section{Related Work}
\label{sec:related}

The efficiency of supervised fine-tuning of LLMs and data augmentation is highlighted in multiple papers \cite{Shen.322025,Li.6122024,Zhou.5182023,Grattafiori.7312024,Wang.12202022}. Wang et al. \cite{Wang.12202022} showed that generating instructions with the same LLM that is also fine-tuned, significantly improves its performance; similar to the creation of the instruction dataset in our work. Zhou et al. showed in \textit{LIMA: Less Is More for Alignment} \cite{Zhou.5182023} that the quality of training data often outweighs quantity, especially for specialized applications. For code generation of IoT software, Shen et al. \cite{Shen.322025} have shown that small, fine-tuned models trained on domain-specific data perform better than generic LLMs in typical IoT coding tasks. Li et al. \cite{Li.6122024} further found that although LLMs have difficulty synthesizing code from input-output examples, it is possible to solve complex programming by example problems after supervised fine-tuning on these kinds of tasks.

Recently, large language models in text forensics have been adapted for author profiling, i.e., with the objective of identifying personal characteristics of the author based on writing styles \cite{cho2024exploring}. In \cite{sharma2025forensicllm}, the Llama 3.1 8B model was fine-tuned for Q\&A on digital forensic research papers and artifacts and conducted a user survey with forensic professionals. Scanlon et al. \cite{scanlon2023chatgpt} evaluated the potential impact of ChatGPT for forensic tasks and concluded that many applications are unsuitable as the evidence is uploaded to the service provider, and also warned against hallucinations of LLMs. A comprehensive survey \cite{yin2025digital} analyzed where LLMs are used in digital forensics and also outlined their risks. In this survey, no other LLM was used as a code generation tool to support the decision-making process for law enforcement. 

An area in digital forensics, where LLMs have been more widely adopted is cyber security \cite{atlam2025llms}, e.g., cyber threat intelligence \cite{perrina2023agir}, vulnerability detection \cite{tamberg2025harnessing} or penetration testing \cite{deng2023pentestgpt}. Recently, Cherif et al. \cite{cherif2025dfir} introduced a new benchmark for evaluating LLMs in cyber security tasks within the context of incident response, which include realistic forensic challenges as well as disk and memory forensics.

To the best of our knowledge, we are the first to use LLMs in the forensic analysis of identification and related documents to generate plausibility checks, rule-based procedures represented in Python, to assess the internal consistency and correctness of data, supporting decision-making in security-sensitive contexts where comprehensibility is essential.

\section{Document Verification System}
\label{sec:background}
The context of this work is a document verification system that is based on the Django framework\footnote{\url{https://www.djangoproject.com/}} and implemented in Python. The system is an AI-assisted processing platform for all types of documents, such as passports, birth certificates or driver's licenses from all over the world, and is used to check documents for forgeries. Scans of printed documents are uploaded to the platform and various AI models are used to help users detect forgeries faster and more reliably.
The entire code base is proprietary software and, therefore, not available online, which is a major challenge for pre-trained and instruction-tuned LLMs when creating executable code in this domain; this is also visible in the following experiments as without fine-tuning there is no available knowledge about this project online and, consequently, not in the training data of open-source models. The complete project consists of 109,824 lines of code and 78 custom database models based on the Django database-abstraction API. Generating executable Django queries for the database is a significant challenge for LLMs, as the code base and model names of the document verification system are in English, but the values stored in the database for filtering and checking properties are partly in German and partly in English. 
The existing plausibility checks are implemented manually in Python and evaluate the correctness and internal consistency of the data. Some checks are quite simple, only checking that the issuing date of the document is not in the future, while others are more complicated. The most complicated check retrieves all documents of the same type from the database and trains a linear regression model to check if the document number correlates with the issuing date. An imaginary check for data integrity is shown in Listing \ref{lst:example-check}, which checks if driver licenses from Germany are of plastic -- the check for ``Kunststoff" in line 16 -- if they are issued in between March 2000 and March 2010.

\begin{python}[numbers=left,frame=tb, caption={Example of a plausibility check, which checks if the material of a given document is of plastic. This check is only executed for German driver's licenses and if the issuing date of the document is between March 2000 and March 2010.},captionpos=b,label=lst:example-check]
def f(document):
    import datetime as dt
    from document_project.models import (DocumentCategory, DocumentElement,DocumentElementEvaluation)
    from plausibility_checks.plausibility_checks import RuleDoesNotApply
    if (document.issuing_country.code != 'DE'
        or document.doc_type.doc_category != DocumentCategory.objects.get(doc_category='Führerschein, national')
        or document.issuing_date is None
        or document.issuing_date < dt.date(year=2000, month=3, day=1)
        or document.issuing_date > dt.date( year=2010, month=3, day=31)):
        raise RuleDoesNotApply("Gegebenes Dokument ist nicht für die Regel relevant.")
    material_evals = DocumentElementEvaluation.objects.filter(document=document,
            check_result__document_element=DocumentElement.objects.get(name='Material'))
    result_dict = {}
    for i, material_eval in enumerate(material_evals):
        result_dict[f'Material {i}'] = material_eval.check_result.check_result_category
    result = material_eval.exclude(check_result__check_result_category__iexact='Kunststoff').count() > 0
    return result, result_dict
\end{python}

There are still a few important constraints we would like to clarify. The document verification system is designed to assess whether documents are authentic or potentially forged. However, the plausibility checks described here represent only a small part of the overall system. They serve merely as indicators of possible inconsistencies or signs of falsification.  A warning triggered by a plausibility check does not necessarily mean that a document is forged; it may still be genuine. In the end, these checks are intended to support experts by providing them with indications of areas that they should review before drawing a conclusion about the document in question.

Furthermore, our system is specifically designed for scanned images of printed documents. Electronic documents are not taken into account, nor are electronic fingerprints of documents or PDF signatures.

\section{Phase One: Python Code Fine-Tuning}
\label{sec:phase1}

\subsection{Fine-Tuning Process}
\label{subsec:phase1_process}

In the first phase, two different fine-tuning strategies derived from the Python code base of the document verification system are evaluated for Llama 3.1 8B \cite{Grattafiori.7312024} and OpenCoder 8B \cite{Huang.1172024}. The primary objective is to ensure that the models develop a fundamental understanding of the code structure, allowing them to comprehend the relation of the Python classes and data flows within the system. The key research question is whether training on raw code base provides a solid foundation for generating more precise plausibility checks in the subsequent phase.

\subsubsection{Objective and Methodology}
\label{ssubsec:phase1_obj}
The main objective of this phase is to embed both semantic and structural understanding of the document verification system code base into the language models. This enables the models to accurately interpret the syntax and semantics of the specific code ecosystem, translate user queries into executable code by correctly leveraging existing functions and libraries, generate relevant code snippets to address technical challenges or analyze particular properties of documents. 

The fine-tuning process consists of several key steps. First, in the data preparation stage, relevant code is extracted, cleaned, and tokenized to construct a diverse and representative training dataset. Next, during model training, the LLMs are fine-tuned using this dataset, with ongoing experimental evaluation and validation to monitor learning progress. Finally, the evaluation phase focuses on verifying the syntactic and semantic correctness of the code generated by the models.

A comprehensive analysis of the impact of this pre-training will be conducted after the completion of the two fine-tuning phases in order to assess its influence on the effectiveness of the model in the subsequent phase for generating plausibility checks.


\subsubsection{Dataset Preparation}
\label{ssubsec:phase1_data}

The training dataset was derived from the document verification system code base, methodically segmented into smaller, logically coherent sections. This segmentation facilitated targeted training and efficient processing by the model. Modules, classes, and functions were identified as central structural elements and separated accordingly. The preprocessing steps included the removal of repetitive comments, code formatting, and tokenization. The resulting dataset consisted of 109,824 lines of code, which were segmented into 310 chunks using a maximum sequence length of 8192 tokens. 

\subsubsection{Instruction Dataset Generation}
\label{ssubsec:phase1_synth}

To enhance the efficiency of fine-tuning smaller models, the more powerful Llama 3.3 70B model was employed to generate synthetic data based on the existing code base. This approach follows the methodology outlined in \cite{Grattafiori.7312024}, where synthetic data generated by a larger model improves the learning quality of smaller models.

The backend code base is used as the basis to generate prompts in the so-called \textit{Instruction-Format} \cite{Wang.12202022} for fine-tuning using the bigger Llama 3.3 70B model. We were able to deploy the 70B model on a single Nvidia RTX 6000 48GPU by using nested 4-bit quantization \cite{dettmers2023qlora}. Each dataset entry follows a uniform structure of task, input, and expected output:

\begin{itemize}
    \item Instruction: Description of the task
    \item Input: Input data or parameters (optional)
    \item Output: Expected code
\end{itemize}

We let the Llama 3.3 model create instructions for the code blocks from the Python code base. If no specific input data is required, the input remains \texttt{None}. The instruction dataset was segmented into 21{,}368 chunks, each with a maximum length of 8192 tokens. This dataset is therefore 700 times the size of the code base dataset. 

\subsection{Experiments}
\label{subsec:phase1_results}
\subsubsection{Evaluation Data} 
To evaluate the fine-tuned models, various coding queries were defined and prototypes of sample solutions were created using Llama 3.3 70B. All prototypes were reviewed and corrected by experts before being used in the evaluation. These queries were categorized into three levels based on their complexity: Low, Medium and High. The complexity was determined by the depth of the queries and the number of nested structures within the code base. The following Table~\ref{tab:queries_code} shows examples of queries and Python code snippets categorized by complexity level.

\setlength{\tabcolsep}{5pt}
\begin{table*}[tb]
    \centering
    \caption{Example queries and generated Python code snippets categorized by complexity level for accessing different document properties.}
    \label{tab:queries_code}
    \renewcommand{\arraystretch}{0.9}
    \begin{tabular}{@{} p{1.5cm} p{4cm} p{11.8cm} @{}}
        \toprule
        \textbf{Complexity}                                                                                                                         & \textbf{Query} & \textbf{Code} \\
        \midrule
        \textbf{Low}                                                                                                                                &
        Generate Python code for the project, that retrieves and prints all barcodes.                                                   & \vspace{-0.5cm}
        \begin{python}[language=Python,breaklines=true]
from example.barcode_detection.models import Barcode
barcodes = Barcode.objects.all()
print(barcodes)
        \end{python}                                                                                       \\
        \textbf{Mid}                                                                                                                                &
        Generate Python code for the project, that retrieves and updates all DocumentElementField objects and prints them out.         &\vspace{-0.5cm}
        \begin{python}[language=Python,breaklines=true]
from example.models import DocumentElementField
fields = DocumentElementField.objects.filter(field_type='example.DateField')
for field in fields:
    field.field_type = 'example.DateField'
    field.save()
    print(field)
        \end{python}                                                                                                \\
        \textbf{High}                                                                                                                               &
        Get all visa requirements for countries that have a specific visa requirement information (e.g., identifier=1) and print them out. &\vspace{-0.5cm}
        \begin{python}[language=Python,breaklines=true]
from example.models import Country, VisaRequirementInformation, VisaRequirement
visa_req_info = VisaRequirementInformation.objects.get(identifier=1)
countries = Country.objects.filter(visarequirementinformation=visa_req_info)
visa_requirements = VisaRequirement.objects.filter(country_of_entry__in=countries)
print(visa_requirements)
        \end{python}                                                                                      \\
        \bottomrule

    \end{tabular}
\end{table*}

\subsubsection{Experimental Setup}
We compare several fine-tuning strategies against baseline models, which were not adapted to the specific task. These baseline models serve as a reference point for assessing performance improvements. Three levels of fine-tuning are considered: (1) self-supervised fine-tuning, on the Python code base from the project; (2) instruction fine-tuning, based on generated instruction dataset derived from the code base; and (3) a combined fine-tuning approach that applies both fine-tuning strategies sequentially. The evaluation is performed using ten queries per complexity level, and each query is executed five times to mitigate the effect of randomness. A generation is successful if it exactly matches the reference implementation.

\subsubsection{Evaluation Metrics}
\label{subsec:metrics}
We employ four different evaluation metrics. The \textit{Success Rate} measures the percentage of responses generated that match the reference solution. Only the output of the programs is used for comparison and only exact matches count as correct solutions. This is similar to calculating the accuracy in classification tasks. The other two metrics are based on the Gestalt-pattern matching algorithm \cite{ratcliff1988pattern}, which calculates the similarity between two string sequences $S_1$ and $S_2$:
\begin{equation}
  Sim(S_1, S_2) = \frac{2M}{|S_1|+|S_2|}
\end{equation}
where $M$ is the number of consecutive matching characters and $|S_1|$ and $|S_2|$ are the length of the string sequences. This metrics complements the Success Rate since it also accounts for partial matches in sequences. We define the \textit{Code Match} to assess the syntactic similarity of the generated program with the reference solution. We decided to check for syntactic similarity, since the custom code base has many specific class names and object values, so this metric indicates if the LLMs have acquired knowledge about the code base of the document verification system. The Code Match calculates the average similarity of all programs with the corresponding reference solution.
The \textit{Output Match} evaluates the similarity between the generated output and the reference output. Similarly to the Code Match, the Output Match is the average of the Gestalt-pattern similarity of all generated outputs for the test cases. 
We also report the \textit{pass@k} metric introduced in \cite{Chen.772021}, which measures if at least one of $k$ generated completions correctly solves a given coding task.

\subsubsection{Results}
Table~\ref{tab:phase_one_finetuning_results} presents the final evaluation of phase 1. There are notable performance differences visible between the models and their respective fine-tuning strategies. OpenCoder consistently achieves high success rates across all complexity levels, whereas Llama shows less consistency, particularly on high-complexity tasks. The OpenCoder Instruct model, fine-tuned with both the instruction dataset and plain code, achieves a 100\% success rate in low- and mid-level queries and 70\% in high-level queries. In contrast, the Llama Instruct model, fine-tuned solely on the instruction dataset, reaches success rates of 54\% for low-level, 64\% for mid-level, and only 20\% for high-level queries. These results suggest that the combined fine-tuning approach used with OpenCoder enhances the model’s ability to generalize across varying levels of complexity and better interpret both structural and semantic aspects of the code.

\begin{table*}[tb]
    \centering
    \caption{Performance of the base models and after various fine-tuning on the code, the instruction dataset or both. L stands for Llama 3.1 8B, OC for OpenCoder 8B, B for the base model and I for the instruction tuned model. We report the Success Rate \textit{SR}, Output Match \textit{OM} and Code Match \textit{CM} }\label{tab:phase_one_finetuning_results}

        \begin{tabular}{lc  ccc  ccc  ccc  ccc}
            \toprule 
                   &         & \multicolumn{3}{c}{\textbf{Baseline}} & \multicolumn{3}{c}{\textbf{Code}} & \multicolumn{3}{c}{\textbf{Instruction Dataset}} & \multicolumn{3}{c}{\textbf{Code \& Instruction}}                        \\
        
            \textbf{Model}  \quad  & \textbf{Level} & \quad \textbf{SR} \quad & \quad \textbf{OM} \quad & \quad \textbf{CM} \quad & \quad \textbf{SR} \quad & \quad \textbf{OM} \quad & \quad \textbf{CM} \quad & \quad \textbf{SR} \quad & \quad \textbf{OM} \quad & \quad \textbf{CM} \quad & \quad \textbf{SR} \quad & \quad \textbf{OM} \quad & \quad \textbf{CM} \quad\\
            
            \cmidrule(){1-2} \cmidrule(lr){3-5} \cmidrule(lr){6-8} \cmidrule(lr){9-11} \cmidrule(l){12-14}
            L-B    & Low     & \cellcolor[rgb]{0.99,0.99,1.00}2 & \cellcolor[rgb]{0.92,0.92,1.00}16 & \cellcolor[rgb]{0.94,0.94,1.00}12 & \cellcolor[rgb]{1.00,1.00,1.00}0 & \cellcolor[rgb]{0.85,0.85,1.00}30 & \cellcolor[rgb]{0.97,0.97,1.00}6 & \cellcolor[rgb]{0.96,0.96,1.00}8 & \cellcolor[rgb]{0.95,0.95,1.00}10 & \cellcolor[rgb]{0.73,0.73,1.00}53 & \cellcolor[rgb]{0.90,0.90,1.00}20 & \cellcolor[rgb]{0.90,0.90,1.00}21 & \cellcolor[rgb]{0.77,0.77,1.00}46 \\
            & Mid     & \cellcolor[rgb]{0.99,0.99,1.00}2 & \cellcolor[rgb]{0.79,0.79,1.00}42 & \cellcolor[rgb]{0.94,0.94,1.00}12 & \cellcolor[rgb]{0.99,0.99,1.00}2 & \cellcolor[rgb]{0.78,0.78,1.00}44 & \cellcolor[rgb]{0.96,0.96,1.00}7 & \cellcolor[rgb]{0.91,0.91,1.00}18 & \cellcolor[rgb]{0.89,0.89,1.00}22 & \cellcolor[rgb]{0.75,0.75,1.00}50 & \cellcolor[rgb]{0.91,0.91,1.00}18 & \cellcolor[rgb]{0.90,0.90,1.00}20 & \cellcolor[rgb]{0.79,0.79,1.00}43 \\
            & High    & \cellcolor[rgb]{1.00,1.00,1.00}0 & \cellcolor[rgb]{0.97,0.97,1.00}5 & \cellcolor[rgb]{0.94,0.94,1.00}12 & \cellcolor[rgb]{1.00,1.00,1.00}0 & \cellcolor[rgb]{0.98,0.98,1.00}4 & \cellcolor[rgb]{0.94,0.94,1.00}12 & \cellcolor[rgb]{0.98,0.98,1.00}4 & \cellcolor[rgb]{0.95,0.95,1.00}10 & \cellcolor[rgb]{0.81,0.81,1.00}38 & \cellcolor[rgb]{0.96,0.96,1.00}8 & \cellcolor[rgb]{0.93,0.93,1.00}15 & \cellcolor[rgb]{0.82,0.82,1.00}36 \\
   
            \cmidrule(){1-2} \cmidrule(lr){3-5} \cmidrule(lr){6-8} \cmidrule(lr){9-11} \cmidrule(l){12-14}
            L-I    & Low     & \cellcolor[rgb]{0.98,0.98,1.00}4 & \cellcolor[rgb]{0.97,0.97,1.00}5 & \cellcolor[rgb]{0.80,0.80,1.00}40 & \cellcolor[rgb]{0.97,0.97,1.00}6 & \cellcolor[rgb]{0.96,0.96,1.00}8 & \cellcolor[rgb]{0.76,0.76,1.00}48 & \cellcolor[rgb]{0.73,0.73,1.00}54 & \cellcolor[rgb]{0.72,0.72,1.00}55 & \cellcolor[rgb]{0.70,0.70,1.00}60 & \cellcolor[rgb]{0.78,0.78,1.00}44 & \cellcolor[rgb]{0.78,0.78,1.00}44 & \cellcolor[rgb]{0.67,0.67,1.00}66 \\
            & Mid     & \cellcolor[rgb]{0.97,0.97,1.00}6 & \cellcolor[rgb]{0.90,0.90,1.00}21 & \cellcolor[rgb]{0.83,0.83,1.00}33 & \cellcolor[rgb]{0.88,0.88,1.00}24 & \cellcolor[rgb]{0.88,0.88,1.00}24 & \cellcolor[rgb]{0.78,0.78,1.00}44 & \cellcolor[rgb]{0.68,0.68,1.00}64 & \cellcolor[rgb]{0.69,0.69,1.00}62 & \cellcolor[rgb]{0.74,0.74,1.00}52 & \cellcolor[rgb]{0.79,0.79,1.00}42 & \cellcolor[rgb]{0.76,0.76,1.00}48 & \cellcolor[rgb]{0.76,0.76,1.00}49 \\
            & High    & \cellcolor[rgb]{0.96,0.96,1.00}8 & \cellcolor[rgb]{0.93,0.93,1.00}15 & \cellcolor[rgb]{0.80,0.80,1.00}41 & \cellcolor[rgb]{0.96,0.96,1.00}8 & \cellcolor[rgb]{0.93,0.93,1.00}14 & \cellcolor[rgb]{0.84,0.84,1.00}32 & \cellcolor[rgb]{0.90,0.90,1.00}20 & \cellcolor[rgb]{0.86,0.86,1.00}28 & \cellcolor[rgb]{0.83,0.83,1.00}33 & \cellcolor[rgb]{0.91,0.91,1.00}18 & \cellcolor[rgb]{0.89,0.89,1.00}23 & \cellcolor[rgb]{0.80,0.80,1.00}39 \\
                   
            \cmidrule(){1-2} \cmidrule(lr){3-5} \cmidrule(lr){6-8} \cmidrule(lr){9-11} \cmidrule(l){12-14}
            OC-B   & Low     & \cellcolor[rgb]{0.98,0.98,1.00}4 & \cellcolor[rgb]{0.97,0.97,1.00}5 & \cellcolor[rgb]{0.81,0.81,1.00}37 & \cellcolor[rgb]{0.85,0.85,1.00}30 & \cellcolor[rgb]{0.85,0.85,1.00}30 & \cellcolor[rgb]{0.77,0.77,1.00}47 & \cellcolor[rgb]{0.60,0.60,1.00}80 & \cellcolor[rgb]{0.59,0.59,1.00}82 & \cellcolor[rgb]{0.62,0.62,1.00}76 & \cellcolor[rgb]{0.60,0.60,1.00}80 & \cellcolor[rgb]{0.55,0.55,1.00}90 & \cellcolor[rgb]{0.65,0.65,1.00}70 \\
            & Mid     & \cellcolor[rgb]{0.97,0.97,1.00}6 & \cellcolor[rgb]{0.95,0.95,1.00}10 & \cellcolor[rgb]{0.81,0.81,1.00}37 & \cellcolor[rgb]{0.90,0.90,1.00}20 & \cellcolor[rgb]{0.75,0.75,1.00}50 & \cellcolor[rgb]{0.80,0.80,1.00}41 & \cellcolor[rgb]{0.85,0.85,1.00}30 & \cellcolor[rgb]{0.84,0.84,1.00}32 & \cellcolor[rgb]{0.81,0.81,1.00}38 & \cellcolor[rgb]{0.70,0.70,1.00}60 & \cellcolor[rgb]{0.69,0.69,1.00}62 & \cellcolor[rgb]{0.70,0.70,1.00}61 \\
            & High    & \cellcolor[rgb]{0.96,0.96,1.00}8 & \cellcolor[rgb]{0.93,0.93,1.00}15 & \cellcolor[rgb]{0.83,0.83,1.00}34 & \cellcolor[rgb]{1.00,1.00,1.00}0 & \cellcolor[rgb]{0.97,0.97,1.00}6 & \cellcolor[rgb]{0.78,0.78,1.00}44 & \cellcolor[rgb]{0.75,0.75,1.00}50 & \cellcolor[rgb]{0.79,0.79,1.00}43 & \cellcolor[rgb]{0.72,0.72,1.00}56 & \cellcolor[rgb]{0.75,0.75,1.00}50 & \cellcolor[rgb]{0.74,0.74,1.00}51 & \cellcolor[rgb]{0.72,0.72,1.00}55 \\
                
            \cmidrule(){1-2} \cmidrule(lr){3-5} \cmidrule(lr){6-8} \cmidrule(lr){9-11} \cmidrule(l){12-14}
            OC-I   & Low     & \cellcolor[rgb]{1.00,1.00,1.00}0 & \cellcolor[rgb]{0.98,0.98,1.00}3 & \cellcolor[rgb]{0.62,0.62,1.00}75 & \cellcolor[rgb]{0.80,0.80,1.00}40 & \cellcolor[rgb]{0.80,0.80,1.00}40 & \cellcolor[rgb]{0.65,0.65,1.00}70 & \cellcolor[rgb]{0.50,0.50,1.00}100 & \cellcolor[rgb]{0.50,0.50,1.00}100 & \cellcolor[rgb]{0.57,0.57,1.00}85 & \cellcolor[rgb]{0.50,0.50,1.00}100 & \cellcolor[rgb]{0.50,0.50,1.00}100 & \cellcolor[rgb]{0.55,0.55,1.00}89 \\
            & Mid     & \cellcolor[rgb]{1.00,1.00,1.00}0 & \cellcolor[rgb]{0.97,0.97,1.00}5 & \cellcolor[rgb]{0.72,0.72,1.00}57 & \cellcolor[rgb]{0.80,0.80,1.00}40 & \cellcolor[rgb]{0.78,0.78,1.00}45 & \cellcolor[rgb]{0.69,0.69,1.00}62 & \cellcolor[rgb]{0.60,0.60,1.00}80 & \cellcolor[rgb]{0.60,0.60,1.00}80 & \cellcolor[rgb]{0.66,0.66,1.00}69 & \cellcolor[rgb]{0.50,0.50,1.00}100 & \cellcolor[rgb]{0.50,0.50,1.00}100 & \cellcolor[rgb]{0.68,0.68,1.00}65 \\
            & High    & \cellcolor[rgb]{0.95,0.95,1.00}10 & \cellcolor[rgb]{0.92,0.92,1.00}16 & \cellcolor[rgb]{0.72,0.72,1.00}55 & \cellcolor[rgb]{0.90,0.90,1.00}20 & \cellcolor[rgb]{0.89,0.89,1.00}23 & \cellcolor[rgb]{0.72,0.72,1.00}57 & \cellcolor[rgb]{0.85,0.85,1.00}30 & \cellcolor[rgb]{0.84,0.84,1.00}31 & \cellcolor[rgb]{0.68,0.68,1.00}65 & \cellcolor[rgb]{0.65,0.65,1.00}70 & \cellcolor[rgb]{0.66,0.66,1.00}68 & \cellcolor[rgb]{0.75,0.75,1.00}50 \\
            \bottomrule                           
        \end{tabular}
    
\end{table*}

The Pass@5 metric indicates whether at least one of five generated completions correctly solves a given coding task (1 if true, else 0). Table~\ref{tab:best_models_passk_compact} compares the best OpenCoder (Instruction Dataset and Code Base Fine-Tuning) and Llama (Instruction Dataset Fine-Tuning) models, which also shows that OpenCoder solves more test cases than Llama.

\begin{table}[tb]
    \centering
    \caption{Pass@5 for check execution – each entry indicates whether $\geq$1 out of 5 solutions passed the test case (T1-T10).}
    \label{tab:best_models_passk_compact}
    \resizebox{\linewidth}{!}{
    \begin{tabular}{l c *{10}{c}}
        \toprule
        \textbf{Model} & \textbf{Level} &  \textbf{T1} & \textbf{T2} & \textbf{T3} & \textbf{T4} & \textbf{T5} & \textbf{T6} & \textbf{T7} & \textbf{T8} & \textbf{T9} & \textbf{T10} \\
        \midrule
        \multirow{3}{*}{Llama}
& Low & \cellcolor{pass}1& 0\cellcolor{fail}& \cellcolor{pass}1& \cellcolor{pass}1& \cellcolor{pass}1& 0\cellcolor{fail}& \cellcolor{pass}1& \cellcolor{pass}1& \cellcolor{pass}1& \cellcolor{pass}1\\ 
& Mid & \cellcolor{pass}1& \cellcolor{pass}1& \cellcolor{pass}1& \cellcolor{pass}1& \cellcolor{pass}1& 0\cellcolor{fail}& \cellcolor{pass}1& \cellcolor{pass}1& 0\cellcolor{fail}& \cellcolor{pass}1\\ 
& High& 0\cellcolor{fail}& \cellcolor{pass}1& 0\cellcolor{fail}& \cellcolor{pass}1& \cellcolor{pass}1& 0\cellcolor{fail}& 0\cellcolor{fail}& \cellcolor{pass}1& \cellcolor{pass}1& 0\cellcolor{fail}\\
        \addlinespace
        \multirow{3}{*}{OpenC.}
& Low & 1 \cellcolor{pass}& 1\cellcolor{pass}& 1\cellcolor{pass}& 1\cellcolor{pass}& 1\cellcolor{pass}& 1\cellcolor{pass}& 1\cellcolor{pass}& 1\cellcolor{pass}& 1\cellcolor{pass}& 1\cellcolor{pass}\\
& Mid & 1 \cellcolor{pass}& 1\cellcolor{pass}& 1\cellcolor{pass}& 1\cellcolor{pass}& 1\cellcolor{pass}& 1\cellcolor{pass}& 1\cellcolor{pass}& 1\cellcolor{pass}& 1\cellcolor{pass}& 1\cellcolor{pass}\\
& High& 1 \cellcolor{pass}& 1\cellcolor{pass}& 0\cellcolor{fail}& 1\cellcolor{pass}& 1\cellcolor{pass}& 1\cellcolor{pass}& 1\cellcolor{pass}& 0\cellcolor{fail}& 1\cellcolor{pass}& 0\cellcolor{fail}\\
        \bottomrule
    \end{tabular}}
\end{table}

The superior performance of OpenCoder can likely be attributed to its pre-training on code-specific datasets, enabling it to capture recurring syntactic patterns and programming patterns more effectively. In contrast, the Llama model, being a general-purpose language model trained on broad textual data, lacks specialized expertise in programming only task. Overall, the OpenCoder Instruct model with combined fine-tuning demonstrates the most reliable and best performance across all evaluation scenarios.

\section{Phase Two: Plausibility Checks}
\label{sec:phase2}

\subsection{Fine-Tuning Process}
\label{subsec:phase2_process}

In the second phase, the LLMs, previously fine-tuned on the document verification system code, are further trained on existing plausibility checks and document properties. The objective is to optimize the models to generate new, technically sound, and relevant plausibility checks. Since both instruct models outperformed their base counterparts, we focus our experiments in this phase on the instruct models. 

\subsubsection{Objective and Methodology}
\label{ssubsec:phase2_obj}

The focus is on developing validation procedures that identify forgeries based on specific document properties. The plausibility checks are defined by functions operating on document objects. The methodology involves fine-tuning the LLM on existing plausibility checks, enabling the generation of functional validation mechanisms from natural language descriptions, such as ``For German driving licenses issued between March 2000 and March 2010, the material must be plastic.'' for Listing \ref{lst:example-check}. This description is paired with input and output examples, i.e. the document as input and the output of the executed check, which is always a Boolean value indicating whether the check was triggered, with a Python dictionary for additional information. The goal is to ensure the model can independently create code for validation mechanisms, considering both structural and content aspects of documents. The final evaluation compares model-generated checks with manually created reference solutions to assess the model's ability to develop precise and technically correct validation rules.

\subsubsection{Dataset Preparation}
\label{ssubsec:phase2_data}

The dataset preparation involves collecting and processing documents for training and evaluation of plausibility checks, including both genuine and forged examples. A key component of the fine-tuning process is 35 expert-created plausibility checks, each extended with 10 input-output examples, totaling 350 examples. These examples ensure the model understands the logic of plausibility checks and can apply them to specific documents.

To improve the generalizability and diversity of the dataset, synthetic plausibility checks were generated, similar to the instruction dataset in phase 1. The Llama 3.3 70B model was used to create new, similar checks based on existing plausibility checks. This method produced 100 additional synthetic plausibility checks, each enriched with specific input-output examples, increasing the total to 450 examples. The dataset comprises 100 generated entries in addition to the 350 existing examples, totaling in 450 chunks (with a chunk length of 8192 tokens).

\subsubsection{Document Properties as Dataset}
\label{ssubsec:phase2_docprops}

A dataset with document-specific properties was created to further extend the model's knowledge. For example, a specific type of identification document may only be issued between 2008 and 2012. This fine-tuning enables the model to process and interpret such context-specific information accurately. An example of such a dataset is shown in Table~\ref{tab:document_properties}. 

\begin{table}[t]
    \centering
    \caption{Example values for properties of a specific document.}
    \label{tab:document_properties}
    \begin{tabular}{llll}
        \toprule
        \textbf{Document ID} & \textbf{Property} &  \textbf{Original Value}  & \textbf{Translation}                   \\
        \midrule
        8713426     & Document Type & Reisepass & Passport             \\
                    & Issuing Country & Deutschland & Germany          \\
                    & Assessment & Fälschung & Forged \\
                    & Document Number & X12345 & \\
                    & Issuing Date & 2020-05-01 & \\
        \bottomrule
    \end{tabular}
\end{table}
These properties help improve the accuracy of model predictions by focusing on specific document characteristics. The document properties dataset comprises 100{,}000 entries, resulting in 100{,}000 chunks (with a chunk length of 8192 tokens). 

\subsection{Experiments}
\label{subsec:phase2_results}

\subsubsection{Evaluation Data}
To prepare the plausibility checks used to assess the performance of the models, specific properties of the provided data were artificially altered. These adjustments are solely for evaluation purposes, testing the models' ability to detect manipulations. Characteristics such as material, security technology, printing technique, or personalization were synthetically modified. These artificial adjustments form the basis for creating plausibility checks formulated in natural language. An example of such a synthetic change is shown in Table~\ref{tab:synthetic_material_change}. For evaluation, 10 plausibility checks per difficulty level (Mid, High) were defined. A check was considered successful if the generated program's output matched the reference output exactly.

\begin{table}[t]
    \centering
    \caption{Synthetic Changes in Document Materials. For clarity we use the English translation of the values.}
    \label{tab:synthetic_material_change}
    \begin{tabular}{lll}
        \toprule
        \textbf{Material} & \textbf{Type} & \textbf{Evaluation}                                                     \\
        \midrule
        Cover             & Plastic \quad     & \cellcolor{red!60}\textcolor{white}{deviating}                          \\

        Page 1            & Paper         & \cellcolor{green!60!black}\textcolor{white}{matches specification} \\
        Page 3            & Paper         & \cellcolor{green!60!black}\textcolor{white}{matches specification} \\
        Page 4            & Paper         & \cellcolor{green!60!black}\textcolor{white}{matches specification} \\
        Page 5            & Paper         & \cellcolor{green!60!black}\textcolor{white}{matches specification} \\
        Page 6/7  \quad        & Paper         & \cellcolor{green!60!black}\textcolor{white}{ matches specification } \\
        \bottomrule
    \end{tabular}
\end{table}

\subsubsection{Experimental Setup}
In phase 2 we use as a baseline the OpenCoder Instruct and Llama Instruct models. These are also the basis for further  fine-tuning on the plausibility checks. To assess the influence of the previous phase, we fine-tune the best OpenCoder and Llama model from phase 1 on the plausibility checks, and also on document properties. 
The same evaluation metrics as in phase 1 are used (see Sect. \ref{subsec:metrics}). Similarly to the previous experiments, each test is repeated five times to mitigate the effects of randomness.  

\paragraph{Pre-experimental Observations:}
A major issue identified during the analysis of the generated plausibility checks is that LLMs often incorrectly generate the following code section:
\begin{python}[language=Python, breaklines=true, frame=tb, captionpos=b, caption={The code section at the beginning of plausibility checks that checks if a given document is relevant for this check.}, label=lst:flawed-check]
from plausibility_checks . plausibility_checks import RuleDoesNotApply
if (document.example != 'example'):
    raise RuleDoesNotApply("Given document is not relevant for this check.")
\end{python}

This checks whether a given document is relevant for the plausibility check. See Listing \ref{lst:example-check} for a more detailed example, which only executes the check if a German driver's license was issued between March 2000 and March 2010, and otherwise throws an exception that the given document is not relevant. 
This code section only occurs in a few plausibility checks, making it difficult for LLMs to learn the correct implementation, even though the rest of the check is implemented correctly. As a result, the error rate of plausibility checks increases significantly, necessitating a more robust evaluation approach. To address this issue, we define two evaluation methods:
\begin{itemize}
    \item \textbf{Exact}: This is a strict evaluation where no modification of the generated program code is made. 
    \item \textbf{Regex}: This evaluation uses a regex expression to modify the code snippet presented in Listing \ref{lst:flawed-check} to print a message instead of raising an exception. With this modification, the output of the check is evaluated without interrupting its execution.
\end{itemize}

\subsubsection{Results}
Table~\ref{tab:phase_two_finetuning_results} summarizes the performance of the models in generating plausibility checks after the second fine-tuning phase. The table compares their performance across two complexity levels (Mid, High) using two evaluation methods, highlighting the impact of domain-specific fine-tuning on the models' ability to generate functionally correct and executable validation routines. The evaluation results indicate that Llama and OpenCoder exhibit different strengths depending on the fine-tuning approach. The combination of fine-tuning the best code model of phase 1 on document properties and plausibility checks leads to significant performance improvements. Llama outperforms OpenCoder, particularly in integrating document properties and fine-tuning on plausibility checks. This is attributed to its broader, generalized pre-training, which enhances its understanding of document structures, since the document properties are in German. This is reasonable since Llama 3.1 is a multilingual model for seven languages, including German. 
OpenCoder also improved its performance through targeted fine-tuning on plausibility checks, but it still lagged behind Llama. Notably, OpenCoder shows stronger performance in the mid-complexity checks.

The impact of different fine-tuning and multistage fine-tuning is particularly visible. Baseline models, without any fine-tuning, demonstrate low success rates for all evaluation metrics. Introducing fine-tuning on plausibility checks alone results in moderate gains in the output match; however, this strategy alone does not produce substantial improvements. In contrast, applying targeted fine-tuning on plausibility checks to the best-performing models from Phase 1 --those already fine-tuned on backend code -- yields significant performance improvements. The most effective approach combines all stages, starting with the best model from the previous phase and all fine-tuning datasets. This comprehensive strategy achieves the highest overall performance, with Llama showing the greatest improvement at this stage due to its enhanced ability to interpret document properties.

\begin{table}[tb]
    \centering
    \caption{Performance of the generated plausibility checks for the base models and after various fine-tuning on the plausibility check dataset (denoted by FT), the document properties (denoted by D) or both. L stands for Llama 3.1 8B, OC for OpenCoder 8B, I for the instruction tuned model and BC for the best code model from phase 1. We report the Success Rate \textit{SR}, Output Match \textit{OM} and Code Match \textit{CM} for the Exact evaluation and the adjusted Regex evaluation.}
    \label{tab:phase_two_finetuning_results}
\resizebox{\linewidth}{!}{
\begin{tabular}{lc ccc ccc}
    \toprule
           &   & \multicolumn{3}{c}{\textbf{Exact}} & \multicolumn{3}{c}{\textbf{Regex}} \\ 
    \textbf{Model} & \textbf{Level} & \quad \textbf{SR} \quad & \quad \textbf{OM} \quad & \quad \textbf{CM} \quad & \quad \textbf{SR} \quad & \quad \textbf{OM} \quad & \quad \textbf{CM} \quad \\ 
    
    \cmidrule(){1-2} \cmidrule(lr){3-5} \cmidrule(lr){6-8} 
    
    L-I          & Low & \cellcolor[rgb]{1.00,1.00,1.00}0 & \cellcolor[rgb]{0.95,0.95,1.00}9 & \cellcolor[rgb]{0.98,0.98,1.00}3 & \cellcolor[rgb]{1.00,1.00,1.00}0 & \cellcolor[rgb]{0.95,0.95,1.00}10 & \cellcolor[rgb]{0.99,0.99,1.00}2 \\ 
    & Mid & \cellcolor[rgb]{1.00,1.00,1.00}0 & \cellcolor[rgb]{0.96,0.96,1.00}8 & \cellcolor[rgb]{0.95,0.95,1.00}9 & \cellcolor[rgb]{1.00,1.00,1.00}0 & \cellcolor[rgb]{0.96,0.96,1.00}7 & \cellcolor[rgb]{0.96,0.96,1.00}7 \\ 
                 
   \cmidrule(){1-2} \cmidrule(lr){3-5} \cmidrule(lr){6-8} 
                         
    L-I FT       & Low & \cellcolor[rgb]{1.00,1.00,1.00}0 & \cellcolor[rgb]{0.82,0.82,1.00}35 & \cellcolor[rgb]{0.99,0.99,1.00}2 & \cellcolor[rgb]{0.98,0.98,1.00}4 & \cellcolor[rgb]{0.84,0.84,1.00}32 & \cellcolor[rgb]{0.97,0.97,1.00}5 \\
    & Mid & \cellcolor[rgb]{0.99,0.99,1.00}2 & \cellcolor[rgb]{0.86,0.86,1.00}27 & \cellcolor[rgb]{0.99,0.99,1.00}1 & \cellcolor[rgb]{0.98,0.98,1.00}4 & \cellcolor[rgb]{0.86,0.86,1.00}28 & \cellcolor[rgb]{0.98,0.98,1.00}4 \\ 
    \cmidrule(){1-2} \cmidrule(lr){3-5} \cmidrule(lr){6-8} 
                         
    L-BC+FT      & Low & \cellcolor[rgb]{0.93,0.93,1.00}14 & \cellcolor[rgb]{0.74,0.74,1.00}51 & \cellcolor[rgb]{0.93,0.93,1.00}14 & \cellcolor[rgb]{0.87,0.87,1.00}26 & \cellcolor[rgb]{0.76,0.76,1.00}49 & \cellcolor[rgb]{0.85,0.85,1.00}30 \\ 
    & Mid & \cellcolor[rgb]{0.89,0.89,1.00}22 & \cellcolor[rgb]{0.82,0.82,1.00}35 & \cellcolor[rgb]{0.86,0.86,1.00}27 & \cellcolor[rgb]{0.83,0.83,1.00}34 & \cellcolor[rgb]{0.82,0.82,1.00}35 & \cellcolor[rgb]{0.82,0.82,1.00}35 \\ 
     \cmidrule(){1-2} \cmidrule(lr){3-5} \cmidrule(lr){6-8} 
                         
    L-BC+D+FT    & Low & \cellcolor[rgb]{0.87,0.87,1.00}26 & \cellcolor[rgb]{0.67,0.67,1.00}67 & \cellcolor[rgb]{0.88,0.88,1.00}25 & \cellcolor[rgb]{0.78,0.78,1.00}44 & \cellcolor[rgb]{0.70,0.70,1.00}60 & \cellcolor[rgb]{0.78,0.78,1.00}44 \\ 
    & Mid & \cellcolor[rgb]{0.90,0.90,1.00}20 & \cellcolor[rgb]{0.79,0.79,1.00}42 & \cellcolor[rgb]{0.91,0.91,1.00}19 & \cellcolor[rgb]{0.84,0.84,1.00}32 & \cellcolor[rgb]{0.79,0.79,1.00}43 & \cellcolor[rgb]{0.84,0.84,1.00}32 \\ 
      \cmidrule(){1-2} \cmidrule(lr){3-5} \cmidrule(lr){6-8} 
                         
    OC-I         & Low & \cellcolor[rgb]{1.00,1.00,1.00}0 & \cellcolor[rgb]{0.95,0.95,1.00}10 & \cellcolor[rgb]{0.99,0.99,1.00}2 & \cellcolor[rgb]{1.00,1.00,1.00}0 & \cellcolor[rgb]{0.95,0.95,1.00}10 & \cellcolor[rgb]{1.00,1.00,1.00}0 \\
    & Mid & \cellcolor[rgb]{1.00,1.00,1.00}0 & \cellcolor[rgb]{0.95,0.95,1.00}9 & \cellcolor[rgb]{0.95,0.95,1.00}10 & \cellcolor[rgb]{1.00,1.00,1.00}0 & \cellcolor[rgb]{0.96,0.96,1.00}7 & \cellcolor[rgb]{1.00,1.00,1.00}0 \\
       \cmidrule(){1-2} \cmidrule(lr){3-5} \cmidrule(lr){6-8} 
                         
    OC-I FT      & Low & \cellcolor[rgb]{1.00,1.00,1.00}0 & \cellcolor[rgb]{0.94,0.94,1.00}11 & \cellcolor[rgb]{1.00,1.00,1.00}0 & \cellcolor[rgb]{1.00,1.00,1.00}0 & \cellcolor[rgb]{0.94,0.94,1.00}11 & \cellcolor[rgb]{1.00,1.00,1.00}0 \\
     & Mid & \cellcolor[rgb]{1.00,1.00,1.00}0 & \cellcolor[rgb]{0.93,0.93,1.00}14 & \cellcolor[rgb]{1.00,1.00,1.00}0 & \cellcolor[rgb]{1.00,1.00,1.00}0 & \cellcolor[rgb]{0.93,0.93,1.00}15 & \cellcolor[rgb]{1.00,1.00,1.00}0 \\
    \cmidrule(){1-2} \cmidrule(lr){3-5} \cmidrule(lr){6-8} 
                         
    OC-BC+FT     & Low & \cellcolor[rgb]{0.95,0.95,1.00}10 & \cellcolor[rgb]{0.74,0.74,1.00}52 & \cellcolor[rgb]{0.95,0.95,1.00}10 & \cellcolor[rgb]{0.90,0.90,1.00}20 & \cellcolor[rgb]{0.74,0.74,1.00}52 & \cellcolor[rgb]{0.89,0.89,1.00}22 \\
    & Mid & \cellcolor[rgb]{0.90,0.90,1.00}20 & \cellcolor[rgb]{0.78,0.78,1.00}45 & \cellcolor[rgb]{0.91,0.91,1.00}19 & \cellcolor[rgb]{0.75,0.75,1.00}50 & \cellcolor[rgb]{0.79,0.79,1.00}43 & \cellcolor[rgb]{0.77,0.77,1.00}46 \\
      \cmidrule(){1-2} \cmidrule(lr){3-5} \cmidrule(lr){6-8} 
                         
    OC-BC+D+FT   & Low & \cellcolor[rgb]{0.95,0.95,1.00}10 & \cellcolor[rgb]{0.78,0.78,1.00}44 & \cellcolor[rgb]{0.94,0.94,1.00}13 & \cellcolor[rgb]{0.90,0.90,1.00}20 & \cellcolor[rgb]{0.75,0.75,1.00}50 & \cellcolor[rgb]{0.91,0.91,1.00}19 \\ 
    & Mid & \cellcolor[rgb]{0.85,0.85,1.00}30 & \cellcolor[rgb]{0.81,0.81,1.00}37 & \cellcolor[rgb]{0.83,0.83,1.00}34 & \cellcolor[rgb]{0.90,0.90,1.00}20 & \cellcolor[rgb]{0.84,0.84,1.00}32 & \cellcolor[rgb]{0.89,0.89,1.00}22 \\ 
    \bottomrule
\end{tabular}}
\end{table}

Table~\ref{tab:best_models_passk_2} shows the pass@5 metric for the best Llama (Best Code, Documents, Checks Finetuned) and OpenCoder (Best Code, Checks Finetuned) models. For low complexity, the Llama model solves more than twice as many checks as OpenCoder. For mid complexity, OpenCoder solves one more check. 

\begin{table}[tb]
    \centering
    \caption{Pass@5 for plausibility check execution – each entry indicates whether $\geq$1 out of 5 solutions passed the test case (C1-C10).}
    \label{tab:best_models_passk_2}
    \resizebox{\linewidth}{!}{
    \begin{tabular}{l c *{10}{c}}
        \toprule
        \textbf{Model} & \textbf{Level} & \textbf{C1} & \textbf{C2} & \textbf{C3} & \textbf{C4} & \textbf{C5} & \textbf{C6} & \textbf{C7} & \textbf{C8} & \textbf{C9} & \textbf{C10} \\
        \midrule
Llama
& Low & 1\cellcolor{pass}& 1\cellcolor{pass}& 0\cellcolor{fail}& 1\cellcolor{pass}& 1\cellcolor{pass}& 1\cellcolor{pass}& 1\cellcolor{pass}& 1\cellcolor{pass}& 1\cellcolor{pass}& 1\cellcolor{pass}\\
& Mid & 1\cellcolor{pass}& 1\cellcolor{pass}& 1\cellcolor{pass}& 1\cellcolor{pass}& 1\cellcolor{pass}& 0\cellcolor{fail}& 1\cellcolor{pass}& 0\cellcolor{fail}& 0\cellcolor{fail}& 0\cellcolor{fail}\\ \addlinespace 
OpenC.      
& Low & 1\cellcolor{pass}& 0\cellcolor{fail}& 1\cellcolor{pass}& 1\cellcolor{pass}& 0\cellcolor{fail}& 0\cellcolor{fail}& 0\cellcolor{fail}& 0\cellcolor{fail}& 0\cellcolor{fail}& 1\cellcolor{pass}\\
& Mid & 0\cellcolor{fail}& 1\cellcolor{pass}& 1\cellcolor{pass}& 1\cellcolor{pass}& 1\cellcolor{pass}& 1\cellcolor{pass}& 0\cellcolor{fail}& 1\cellcolor{pass}& 0\cellcolor{fail}& 1\cellcolor{pass}\\ 
    \bottomrule
    \end{tabular}}
\end{table}

\section{Discussion, Limitations \& Future Work}
\label{sec:discussion}

This study explored the adaptation of pre-trained large language models for generating document verification mechanisms based on Python code on constrained hardware resources. The results demonstrate that fine-tuning LLMs on domain-specific datasets significantly enhances their ability to generate executable validation rules. 

The fine-tuning process revealed distinct strengths in each model. OpenCoder performed exceptionally well in tasks involving code-specific processing, while Llama 3.1 demonstrated superior generalization capabilities, especially in generating plausibility checks where document properties in German need to be combined with the database queries. The multi-stage fine-tuning strategy, initially focused on the project code and later on actual plausibility checks, significantly improved the models' performance compared to non-finetuned baseline models. This multistage approach enhanced the models' ability to generate precise and executable plausibility checks, making them more effective for document verification.

Nevertheless, working with LLM is a challenge, especially due to its black-box nature, which makes it difficult to evaluate the efficiency of certain adjustments or data preparation. Although evaluation metrics can help to measure efficiency, the process is time-consuming. Hardware limitations restricted fine-tuning to smaller 8B models and prevented testing larger models that could provide better results. In addition, handling sensitive data in a complex project environment required caution, but also provided an opportunity to develop innovative solutions. Another problem of LLMs is the need for retraining, when new information about document forgeries is added or the code base of the document verification system is updated. Several promising directions exist for future research. Retrieval-Augmented Generation \cite{lewis2020retrieval} could allow models to dynamically incorporate external, up-to-date knowledge bases (e.g., known forgery patterns or updates of document security features), potentially enhancing validation accuracy without constant retraining. 

Other promising research directions are the integration of LLMs with computer vision methods to enable analysis of visual document features (layout, seals, fonts) for more robust forgery detection. Deploying these models in real-time systems within relevant operational environments (e.g., law enforcement or border control) would provide critical insights into practical performance, scalability, and usability. 

\section{Conclusion}
\label{sec:conclusion}

The findings of this study demonstrate the potential of fine-tuned LLMs for generating programmatic rules for document forgery detection. A multistage supervised fine-tuning approach, initially focusing on the document verification system code and subsequently on specific plausibility checks and document properties, proved effective in adapting the Llama 3.1 and OpenCoder models to this domain-specific task. The results show that fine-tuned models can generate executable and relevant validation code, significantly outperforming the baseline models. Llama models showed better generalization, particularly when fine-tuned on document properties, while OpenCoder excelled after code-specific fine-tuning.

This work lays the foundation for automated document verification systems. The success of fine-tuning confirms that models can generate both code fragments and complete validation routines with reasonable accuracy, offering a scalable tool to improve security processes.




\section*{Acknowledgment}
We extend our gratitude to the Document Forensics Department of the Baden Württemberg State Office of Criminal Investigations for their invaluable assistance in providing the data of the documents essential to this project. Special recognition is owed to Rolf Fauser, whose personal dedication and insightful expertise were instrumental in facilitating our work.

\bibliographystyle{splncs04} 
\bibliography{main.bib} 

\end{document}